\documentclass[10pt,twocolumn,letterpaper]{article}

\usepackage{main}
\usepackage{amsthm}
\usepackage{algorithm}
\usepackage{algpseudocode}
\usepackage{multirow}

\newtheoremstyle{paragraph}
  {0pt}{0pt}{\itshape}{}{\bfseries}{.}{5pt plus 1pt minus 1pt}{}
\theoremstyle{paragraph}
\newtheorem{lemma}{Lemma}
\makeatletter
\newcommand{\spm@runin}[1]{\par\normalfont\topsep\z@\partopsep\z@\relax
  \trivlist\item[\hskip\labelsep\itshape #1\@addpunct{.}]\ignorespaces}
\renewenvironment{proof}[1][\proofname]
  {\pushQED{\qed}\spm@runin{#1}}
  {\popQED\endtrivlist\@endpefalse}
\makeatother

\newcommand{\method}{SPM\xspace}
\newcommand{\E}{\mathbb{E}}
\newcommand{\KL}{\mathrm{KL}}
\newcommand{\Cov}{\mathrm{Cov}}
\newcommand{\sg}{\operatorname{sg}}
\newcommand{\N}{\mathcal{N}}
\DeclareMathOperator*{\argmax}{arg\,max}
\newcommand{\cmark}{\ensuremath{\checkmark}}
\newcommand{\xmark}{\ensuremath{\times}}

\definecolor{linkblue}{rgb}{0.21,0.49,0.74}
\usepackage[pagebackref,breaklinks,colorlinks,allcolors=linkblue]{hyperref}
\usepackage[capitalize]{cleveref}

\title{Reinforcing Few-Step Diffusion Models via Shadow Posterior Matching}
\author{
  Jingqiao Xiu\\
  {\tt\small jingqiao.xiu@eng.ox.ac.uk}
  \and
  Yicong Li\\
  {\tt\small liyicong@ustc.edu.cn}
}

\begin{document}
\maketitle
\begin{abstract}
Few-step diffusion models are widely deployed for their efficiency. However, reinforcing them with black-box rewards, as is routine for multi-step models, faces two structural challenges. First, a few-step generator is a sampler rather than a score model, so its velocity is not the forward velocity of the generated distribution. Second, it must be anchored to the undistilled teacher, which shares no transitions with it. Existing few-step RL methods handle the latter with a marginal KL to the teacher, but each pays for the former in its reward credit. We propose Shadow Posterior Matching (\method), which renoises generated samples into shadow inputs and obtains the credit from their shadow posterior. From the KL-regularized reward objective, we derive a regression loss that turns reward values into a reward direction while anchoring the generator's outputs to the teacher and aligning the shadow inputs with the deployed inputs. Applied to 4-step generators, \method achieves higher rewards with less reward hacking than prior few-step RL methods.
\end{abstract}

\section{Introduction}
\label{sec:intro}

\begin{table*}[t]
\captionsetup{skip=6pt}
\centering
\small
\setlength{\tabcolsep}{4pt}
\renewcommand{\arraystretch}{1.1}
\begin{tabular}{@{}lccccc@{}}
\toprule
\multirow{2}{*}[-0.5ex]{\textbf{Paradigm}} & \multicolumn{4}{c}{\textbf{Reward credit}} & \textbf{Anchor} \\
\cmidrule(lr){2-5}\cmidrule(l){6-6}
& \textbf{Reward value} & \textbf{Final call} & \textbf{Sharp sample} & \textbf{Reward direction} & \textbf{Teacher KL} \\
\midrule
Reward backpropagation~\cite{jiang2026distribution,luo2024diffinstruct,wu2026diffinstruct} & \xmark & -- & \cmark & \cmark & \cmark \\
Policy gradients~\cite{huang2026reinforcing} & \cmark & \xmark & \cmark & \cmark & \cmark \\
Learned surrogate~\cite{luo2026tdmr1} & \xmark & $\circ$ & \cmark & \cmark & \cmark \\
Forward-process residual~\cite{dong2026guiding} & \cmark & \cmark & \xmark & \cmark & \cmark \\
Reward-weighted matching~\cite{jiang2026distribution,lu2025reward,bai2026optimizing} & \cmark & \cmark & \cmark & \xmark & \cmark \\
\midrule
Shadow posterior matching (Ours) & \cmark & \cmark & \cmark & \cmark & \cmark \\
\bottomrule
\end{tabular}
\caption{\textbf{Few-step RL paradigms, distinguished by their reward credit.} \emph{Reward value}: the credit is computed from measured reward values, without the reward's gradient or a learned proxy. \emph{Final call}: the deterministic final call receives credit from a black-box reward ($\circ$ means only through a learned proxy, and -- means not applicable, since the paradigm needs the reward's gradient). \emph{Sharp sample}: no reference in the update is built from the generator's own velocity, so the samples receive no shift toward the posterior mean and stay sharp. \emph{Reward direction}: the reward provides an update direction independent of the anchor's force instead of only reweighting the anchor. \emph{Teacher KL}: the generator is anchored to the teacher by a KL between their marginal distributions.}
\label{tab:paradigms}
\vspace{-4pt}
\end{table*}

Few-step diffusion models trained by distribution matching distillation (DMD)~\cite{yin2024onestep,yin2024improved} sample in a few network calls at nearly the quality of their multi-step teachers. They also inherit their teachers' failures: garbled text, wrong object counts, and broken spatial relations survive distillation. The evaluators that detect these failures, such as OCR engines, object detectors, and vision-language judges, usually return scalar values without usable gradients. For multi-step diffusion models, reinforcement learning (RL) with such black-box rewards is well established, through policy gradients on the reverse process~\cite{black2024training,liu2025flowgrpo,xue2025dancegrpo} or reward-weighted objectives on the forward process~\cite{xue2026advantage,zheng2026diffusionnft,bergmeister2026reinforce}. Applied to a few-step generator, the same algorithms blur its samples~\cite{jia2025reward,luo2026tdmr1}, leave its final call without reward credit~\cite{jia2025reward,huang2026reinforcing}, or let it exploit the reward~\cite{ren2024hypersd,jiang2026distribution}.

\begin{figure}[t]
\captionsetup{skip=4pt}
\centering
\includegraphics[width=\linewidth]{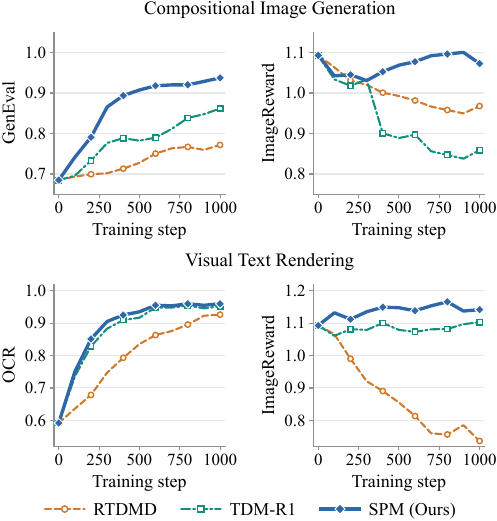}
\caption{\textbf{Reward curves with a single black-box reward.} The reward is GenEval (top) or OCR (bottom). \textbf{Left:} training reward on test prompts. \textbf{Right:} ImageReward on DrawBench prompts. \method achieves higher task rewards with better image quality.}
\label{fig:teaser}
\vspace{-8pt}
\end{figure}

These failures are structural and stem from two challenges. \textbf{(1) Velocity mismatch.} Multi-step RL relies on the policy being a score model, whose velocity is the forward velocity of the generated distribution, so its predictions are posterior means. Forward-process methods use this velocity as the reference of a reward-weighted flow-matching residual~\cite{zheng2026diffusionnft}, and reverse-process methods use it to construct the score of a marginal-preserving stochastic differential equation (SDE)~\cite{liu2025flowgrpo}. A few-step generator is instead a sampler: its few deterministic calls, separated by renoising, output sharp samples rather than posterior means, so its velocity is not the forward velocity of its distribution. The forward-process residual built on this velocity therefore adds a reward-free shift toward the posterior mean, blurring the samples~\cite{luo2026tdmr1}. Likewise, the constructed SDE is not a stochastic version of the sampler. Reverse-process credit can therefore act only through the sampler's renoising kernels, whose large noise gives high-variance gradients~\cite{jia2025reward}, and it never reaches the final call~\cite{jia2025reward,huang2026reinforcing}. \textbf{(2) External anchor.} Multi-step RL keeps the policy close to a reference of the same kind on the same sampler, either by a Kullback--Leibler (KL) divergence between the small Gaussian transitions of the two models~\cite{liu2025flowgrpo} or through a slowly updated old copy~\cite{zheng2026diffusionnft}. For a few-step generator, such a reference is a distilled copy that has already lost teacher modes, and regularizing toward it does not prevent reward exploitation~\cite{ren2024hypersd,jiang2026distribution}. The anchor must be the undistilled teacher, a different kind of model sharing no transitions with the generator. A transition KL therefore cannot be formed, and the deterministic final call has no transition density at all. The only remaining teacher anchor is a marginal KL between their sample distributions.

Most recent few-step RL methods realize this anchor with DMD~\cite{yin2024onestep,yin2024improved}, which pushes the generator's samples along the difference between the teacher score and a fake score trained on those samples. They differ in the reward credit, and each pays a cost for the velocity mismatch. Reward backpropagation~\cite{jiang2026distribution,luo2024diffinstruct,wu2026diffinstruct} reaches the final call by differentiating the reward, so it needs the reward's gradient and cannot handle black-box rewards. Policy gradients~\cite{huang2026reinforcing} admit a black-box reward, but the final call is trained by the anchor alone. A learned surrogate~\cite{luo2026tdmr1} turns the black-box reward into a differentiable proxy, so the generator climbs the gradient of a fit and can exploit its errors. A forward-process residual~\cite{dong2026guiding} retains the shift toward the posterior mean. Reward-weighted matching~\cite{jiang2026distribution,lu2025reward,bai2026optimizing} avoids the generator's velocity by weighting the anchor's force with the reward, but the reward then only rescales this force instead of providing an independent direction. This weighted force therefore still vanishes at the teacher, which remains a rest point~\cite{jiang2026distribution}. Reweighting the fake score's training samples as well moves this rest point, but requires the fake score to model a reward-reweighted distribution, which is harder to learn than the generator's own~\cite{bai2026optimizing}.

A reward credit that avoids all of these costs must be computed from reward values alone, reach every call, take its reference from the generator's distribution rather than its velocity, and provide an independent reward direction. Reaching every call is the hardest, since the output of the final call is fixed by its input. To address this, we propose \textbf{Shadow Posterior Matching (\method)}, which obtains such a credit by renoising the generated sample with fresh noise into an input for any call. Given this input, the generated sample is itself a draw from the posterior of the generator's distribution. We term such an input a \emph{shadow input} of the call and this posterior its \emph{shadow posterior}. Unlike the call's output, the shadow posterior is non-degenerate at every call, so a black-box reward can be correlated with its samples.

Applying the KL-regularized reward objective to each call on its shadow inputs, we derive a regression loss with the following three terms. \textbf{(1) Shadow credit} turns black-box reward values into a reward direction for the call's output through the shadow posterior, with a fake score instead of the generator's velocity as the reference. \textbf{(2) Teacher anchor} keeps the distribution of the call's outputs close to the teacher's. \textbf{(3) Endpoint matching} keeps that distribution close to the generated distribution, so that the shadow inputs of every call match its deployed inputs. Besides the generator, the three terms need only the teacher and the fake score trained by standard flow matching. As summarized in \Cref{tab:paradigms}, our shadow credit meets all requirements, whereas each existing paradigm misses at least one.

Our contributions can be summarized as:
\begin{itemize}[leftmargin=\parindent, itemsep=3pt]
\item We identify two challenges of few-step RL and show that existing paradigms all meet the external anchor with a marginal KL to the teacher but pay for the velocity mismatch in their reward credit.
\item We propose \method, which credits every call of a few-step generator through the shadow posterior of generated samples, with a regression loss derived from the KL-regularized reward objective.
\item We conduct experiments on 4-step DMD-distilled generators, showing that \method achieves higher rewards with less reward hacking than prior few-step RL methods.
\end{itemize}

\section{Related Work}
\label{sec:rel}

\noindent\textbf{Few-step diffusion distillation.} Few-step generators are distilled from a multi-step teacher in two ways. Trajectory-based methods reproduce the teacher's ordinary differential equation (ODE) map, by halving the number of steps repeatedly~\cite{salimans2022progressive} or by enforcing self-consistency along the trajectory~\cite{song2023consistency,luo2023latent,lu2025simplifying}. Distribution-based methods match the student's output distribution to the teacher's, with an adversarial discriminator~\cite{sauer2024adversarial,sauer2024fast} or with a reverse KL whose gradient is the difference between the teacher's score and the score of the student's distribution~\cite{wang2023prolificdreamer,luo2023diffinstruct,yin2024onestep}. Among them, DMD~\cite{yin2024onestep} estimates the student's score online with a fake score trained by denoising on the student's samples, DMD2~\cite{yin2024improved} turns the one-step generator into a few-step sampler that renoises its own predictions between calls, and TDM~\cite{luo2025learning} matches distributions along the student's trajectory. We build on DMD2, reinforcing its generator and adding a per-call condition to its fake score.

\vspace{3pt}
\noindent\textbf{Multi-step diffusion RL.} Reward fine-tuning methods for multi-step models differ in how the reward reaches the parameters. Reward backpropagation differentiates a reward model through the sampler~\cite{xu2023imagereward,clark2024directly,prabhudesai2023aligning,domingoenrich2025adjoint} and needs differentiable rewards. Reverse-process policy gradients treat denoising as a Markov decision process with Gaussian transitions of known density and admit black-box rewards: DDPO~\cite{black2024training} and DPOK~\cite{fan2023dpok} apply PPO-style updates, while Flow-GRPO~\cite{liu2025flowgrpo} and DanceGRPO~\cite{xue2025dancegrpo} combine the group-relative advantages of GRPO~\cite{shao2024deepseekmath} with an ODE-to-SDE conversion that preserves the model's marginals. Forward-process methods avoid trajectory likelihoods by weighting or contrasting denoising and flow-matching losses on clean samples, as in reward-weighted regression~\cite{lee2023aligning,fan2025online}, Diffusion-DPO~\cite{wallace2024diffusion}, AWM~\cite{xue2026advantage}, DiffusionNFT~\cite{zheng2026diffusionnft}, and RAM~\cite{bergmeister2026reinforce}. Both black-box families assume a score-model policy and a reference of the same kind, neither of which holds for a few-step generator.

\vspace{3pt}
\noindent\textbf{Few-step diffusion RL.} Early methods fine-tune consistency models by backpropagating differentiable rewards~\cite{ren2024hypersd} or applying PPO to renoising kernels~\cite{oertell2024rl}. Recent work primarily targets generators distilled via distribution matching~\cite{luo2023diffinstruct,yin2024onestep,luo2025learning}. DI++~\cite{luo2024diffinstruct}, DMDR~\cite{jiang2026distribution}, and DIDR~\cite{wu2026diffinstruct} backpropagate the reward's gradient, which a black-box reward does not provide. RTDMD~\cite{huang2026reinforcing} combines such backpropagation at the final call with Flow-GRPO over the renoising kernels. TDM-R1~\cite{luo2026tdmr1} climbs the gradient of a differentiable surrogate fitted online to the black-box reward. GDMD~\cite{dong2026guiding} adds a DiffusionNFT loss on DMD's update targets with the generator's old velocity as the reference. Re-DMD~\cite{lu2025reward} and AMD~\cite{bai2026optimizing} weight DMD's force by the reward. Like these methods, \method keeps DMD as the teacher anchor, but takes its reward credit from the shadow posterior of generated samples.

\section{Shadow Posterior Matching}
\label{sec:method}

\subsection{Background}
\label{sec:background}

\noindent\textbf{Few-step generator.} In flow matching, a sample $x$ of a distribution $p$ is noised to $y_t=(1-t)x+t\epsilon$ with Gaussian noise $\epsilon\sim\N(0,I)$. The time $t\in[0,1]$ runs from data ($t=0$) to noise ($t=1$). The noised sample $y_t$ has law $p_t$ and moves with velocity $u=\epsilon-x$. We describe $p$ by its \emph{forward velocity} $v_p(y,t)=\E_p[u\mid y_t=y]$, which is the minimizer over fields $v$ of the flow-matching loss $\E\|v(y_t,t)-u\|^2$. Since $\E[\epsilon\mid y_t]=y_t+(1-t)v_p$, the score of $p_t$ is:
\begin{equation}
\nabla_y\log p_t(y)=-\frac{y+(1-t)\,v_p(y,t)}{t}.
\label{eq:score}
\end{equation}

DMD~\cite{yin2024onestep,yin2024improved} distills a pretrained teacher $p_T$ with velocity $v_T=v_{p_T}$ into a student that samples in a few network calls. The student is a velocity network $v_\theta(y,s)$ with parameters $\theta$. It is deployed at $K$ noise levels, or \emph{stops}, $1=s_1>s_2>\dots>s_K>0$. A call at stop $s$ maps its input $y$ to the prediction $\hat x_\theta(y,s)=y-s\,v_\theta(y,s)$. The sampler alternates calls and renoising~\cite{yin2024improved}:
\begin{equation}
\begin{aligned}
z_1&=\epsilon_0,\quad \hat x_j=\hat x_\theta(z_j,s_j),\\
z_{j+1}&=(1-s_{j+1})\hat x_j+s_{j+1}\xi_j,
\end{aligned}
\label{eq:sampler}
\end{equation}
where $z_j$ is the input of the $j$-th call and $\epsilon_0,\xi_j\sim\N(0,I)$. The final sample $x=\hat x_K$ has law $q=q_\theta$. DMD trains the student to minimize the noised divergence $D_\nu(q\|p_T)=\E_{t\sim\nu}\KL(q_t\|p_{T,t})$ for a distribution $\nu$ on $(0,1)$, which vanishes only at $q=p_T$. By \Cref{eq:score}, $\nabla_y\log(\frac{q_t}{p_{T,t}})=\frac{1-t}{t}(v_T-v_q)$. DMD replaces the unknown $v_q$ by a fake score $F\approx v_q$, trained by flow matching on samples of $q$. With $\partial_\theta$ the Jacobian with respect to $\theta$, DMD descends:
\begin{equation}
\nabla_\theta\KL(q_t\|p_{T,t})\approx\tfrac{1-t}{t}\,\E\big[(v_T-F)(y_t)\cdot\partial_\theta y_t\big].
\label{eq:dmd}
\end{equation}

We fine-tune the distilled generator with a scalar reward $R(x)$, accessible only through evaluations, to maximize:
\begin{equation}
\begin{aligned}
J(q)&=\E_q[R]-\alpha\,\KL(q\,\|\,p_T),\\
p^*&=\argmax_q J\propto p_T\,e^{R/\alpha},
\end{aligned}
\label{eq:J}
\end{equation}
where the tilting temperature $\alpha>0$ weights the KL. A lower $\alpha$ tilts $p^*$ more strongly toward high reward. Ascending $J$ requires an anchor for the KL term and a credit for the reward term. Few-step RL methods anchor with the marginal KL $D_\nu(q\|p_T)$ and differ in the reward credit. We formalize the reward-credit columns of \Cref{tab:paradigms} below.

\vspace{3pt}
\noindent\textbf{Reward credit.} The exact gradient of the reward term differentiates the Gaussian renoising kernels in likelihood-ratio form and the deterministic final call pathwise:
\begin{equation}
\begin{aligned}
\nabla_\theta\E[R]={}&\sum_{j=1}^{K-1}\E\Big[R\,\tfrac{1-s_{j+1}}{s_{j+1}}\,\xi_j\cdot\partial_\theta\hat x_\theta(z_j,s_j)\Big]\\
&+\E\big[\nabla R(x)\cdot\partial_\theta\hat x_\theta(z_K,s_K)\big].
\end{aligned}
\label{eq:pg}
\end{equation}
The likelihood-ratio sum needs only reward values. However, the final call enters only the pathwise term, which needs the reward gradient $\nabla R$ or a learned proxy. To credit every call from reward values, one can pull the generator's velocity $v_\theta$ at a renoised final sample $y_s$ along a residual $w\,(u-v_{\mathrm{ref}})$. Here $w$ is a reward-derived weight and $v_{\mathrm{ref}}$ a reference velocity. Given $y_s=y$, the residual has mean:
\begin{equation}
\begin{aligned}
\E\big[w\,(u-v_{\mathrm{ref}})\mid y\big]={}&\Cov_q(w,u\mid y)\\
&+\E[w\mid y]\,\big(v_q-v_{\mathrm{ref}}\big)(y).
\end{aligned}
\label{eq:consistency}
\end{equation}
Only the covariance term carries a reward direction. The second term must therefore vanish, which for a general weight requires $v_{\mathrm{ref}}=v_q$. With $v_\theta$ as the reference, this term would instead move the sharp prediction $\hat x_\theta$ by ${\E[w\mid y]}\,(\E_q[x\mid y_s=y]-\hat x_\theta)$, since $\E_q[x\mid y_s=y]=y-s\,v_q$. This shift lies along the direction to the posterior mean and carries no reward information. Alternatively, the anchor's force $v_T-F$ can replace $u-v_{\mathrm{ref}}$ in the residual. As this force is fixed given $y$, the covariance term disappears. For $F=v_q$, the weighted force has mean:
\begin{equation}
\E\big[w\,(v_T-F)\mid y\big]=\E[w\mid y]\,\big(v_T-v_q\big)(y).
\label{eq:rwdm}
\end{equation}
It vanishes at every $y$ when $q=p_T$. The teacher thus remains a rest point, and the reward provides no direction independent of the anchor's force. A credit from reward values that reaches every call must therefore take $v_q$ as its reference and keep the covariance term of \Cref{eq:consistency}.

\subsection{Method}
\label{sec:spm}
\begin{figure*}[t]
\captionsetup{skip=4pt}
\centering
\includegraphics[width=\textwidth]{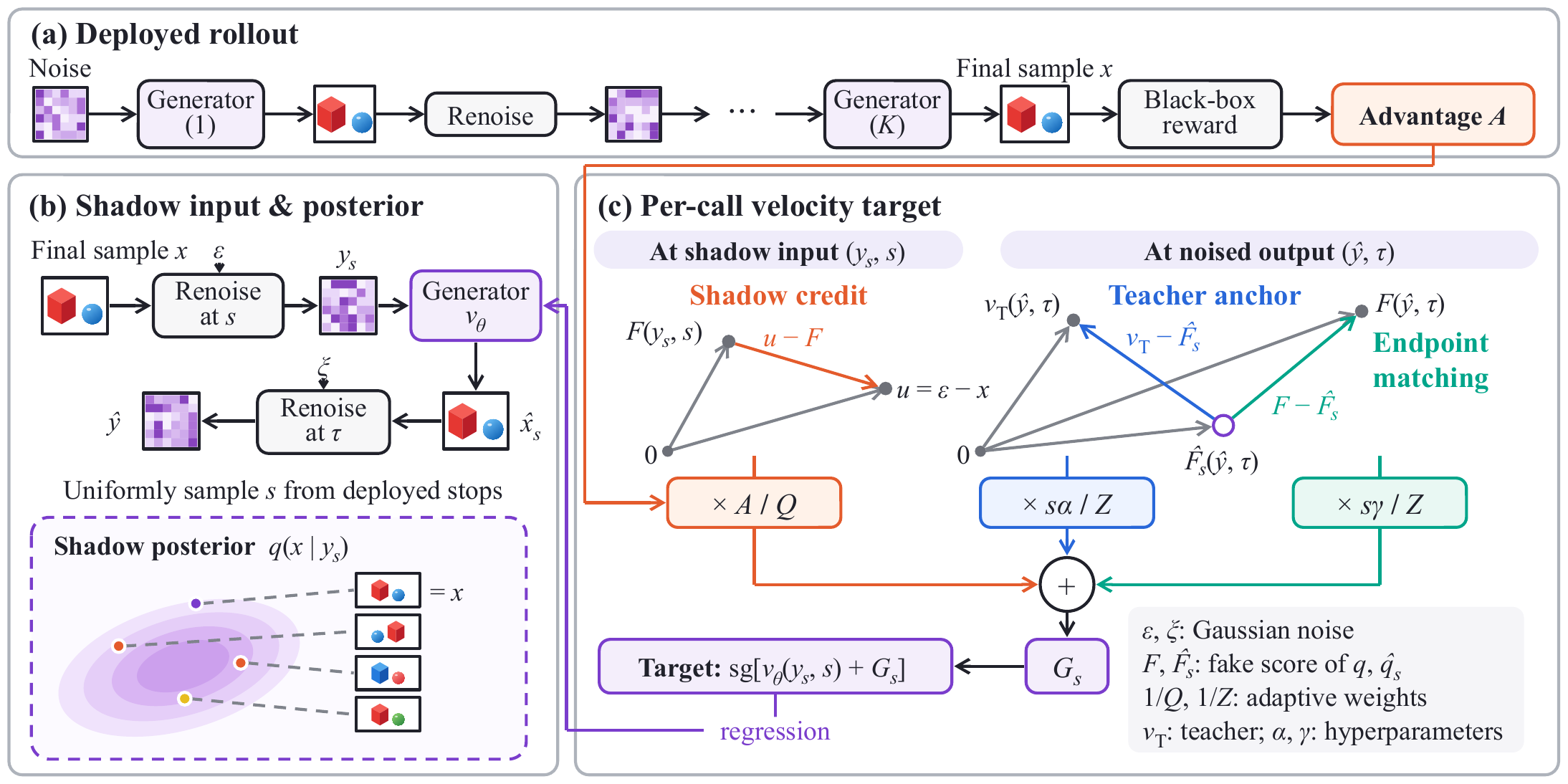}
\caption{\textbf{Overview of Shadow Posterior Matching.} \textbf{(a)} The deployed few-step sampler generates final samples without gradients, and their black-box rewards yield group-relative advantages. \textbf{(b)} Renoising with fresh noise pairs each shadow input $y_s$ with a posterior sample $x$ and the velocity $u$. \textbf{(c)} Colored arrows show three velocity differences: $u-F(y_s,s)$ for shadow credit, and $v_T-\hat F_s$ and $F-\hat F_s$ at $(\hat y,\tau)$ for teacher anchor and endpoint matching. A regression loss pulls the generator's velocity $v_\theta(y_s,s)$ along their weighted sum $G_s$.}
\label{fig:method}
\vspace{-8pt}
\end{figure*}

\noindent\textbf{Shadow inputs.} Even with $v_q$ as the reference, the residual of \Cref{eq:consistency} trains the generator's velocity at $y_s$ as a forward velocity. Its covariance term thus tilts the posterior mean $\E_q[x\mid y_s]$ toward high reward, but the generator predicts a sharp sample, not this mean. \method therefore keeps $y_s$ but treats it as an input of the call at stop $s$, standing in for its deployed input. We term $y_s$ a \emph{shadow input} of the call, and its posterior $q(x\mid y_s)$ the \emph{shadow posterior}. The prediction at a shadow input is the call's output $\hat x_s=\hat x_\theta(y_s,s)$, whose law we write as $\hat q_s$. Since the final sample $x$ is a draw from the shadow posterior, the covariance term over such draws credits this output, not a posterior mean. The stand-in itself is exact if and only if $\hat q_s=q$ at every non-final stop.

\vspace{3pt}
\begin{lemma}[Shadow inputs]
\label{lem:induction}
The deployed input $z_k$ has the law of the shadow input $y_{s_k}$ for every $k$ if and only if $\hat q_{s_j}=q$ for all $j<K$. In that case, also $\hat q_{s_K}=q$.
\end{lemma}
\vspace{3pt}
\begin{proof}
$z_1=\epsilon_0$ and $y_{s_1}=\epsilon$ are standard Gaussian. If $z_k$ and $y_{s_k}$ share a law, so do their outputs. For $k<K$, if this law $\hat q_{s_k}$ is $q$, renoising with fresh noise gives $z_{k+1}$ the law of $y_{s_{k+1}}$. For $k=K$ the output on $z_K$ is $x$, so $\hat q_{s_K}=q$. Conversely, let the inputs match at $s_j$ and $s_{j+1}$. Renoising then maps $\hat q_{s_j}$ and $q$ to the equal laws of $z_{j+1}$ and $y_{s_{j+1}}$. Since $s_{j+1}<1$, this map is injective, so $\hat q_{s_j}=q$.
\end{proof}

\noindent\textbf{Per-call objective.} For every stop, \method applies $J$ to the law of the call's outputs on shadow inputs and additionally penalizes the violation of the condition of \Cref{lem:induction}:
\begin{equation}
J_s(\theta)=\E\big[R(\hat x_s)\big]-\alpha\,D_\nu(\hat q_s\|p_T)-\gamma\,D_\nu(\hat q_s\|q),
\label{eq:Js}
\end{equation}
where $\gamma$ weights the last term. We call the three terms \emph{shadow credit}, \emph{teacher anchor}, and \emph{endpoint matching}. When $\hat q_s=q$, $J_s$ is $J$ with its KL replaced by $D_\nu$. Although $q$ also depends on $\theta$, endpoint matching treats it as fixed and moves the call's outputs toward it. Yet the condition of \Cref{lem:induction} constrains only the first $K-1$ calls and implies $\hat q_{s_K}=q$. Even if it holds only approximately, their endpoint-matching divergences still bound the total-variation gap between $\hat q_{s_K}$ and $q$, the final call's output laws on shadow and deployed inputs. We thus drop endpoint matching at the final stop.

\vspace{3pt}
\noindent\textbf{Per-call gradient.} With $q$ fixed, the law of the shadow inputs is fixed as well. Each term of \Cref{eq:Js} then depends on $\theta$ only through the call's output $\hat x_s$. The two divergences depend on $\hat x_s$ through the noised output $\hat y=(1-\tau)\hat x_s+\tau\xi$, where $\tau\sim\nu$ and $\xi\sim\N(0,I)$. For a reference $p\in\{p_T,q\}$, we apply \Cref{eq:dmd} to $\hat q_s$, with $v_p$ and $v_{\hat q_s}$ in place of $v_T$ and $F$. With $\partial_\theta\hat y=(1-\tau)\,\partial_\theta\hat x_s$, $\kappa(\tau)=(1-\tau)^2/\tau$, and $\partial_\theta\hat x_s=-s\,\partial_\theta v_\theta(y_s,s)$, the divergence gradients and the gradient of $J_s$ are:
{\medmuskip=2mu\thickmuskip=4mu
\begin{equation}
\begin{aligned}
&\nabla_\theta D_\nu(\hat q_s\|p)=\E\big[\kappa(\tau)\,(v_p-v_{\hat q_s})(\hat y,\tau)\cdot\partial_\theta\hat x_s\big],\\
&\nabla_\theta J_s=\E\big[G_s\cdot\partial_\theta v_\theta(y_s,s)\big],\quad G_s=G_s^{\mathrm R}+G_s^{\mathrm D},\\
&G_s^{\mathrm R}=-s\,\nabla R(\hat x_s),\\
&G_s^{\mathrm D}=s\,\kappa(\tau)\big[\alpha(v_T-v_{\hat q_s})+\gamma(v_q-v_{\hat q_s})\big](\hat y,\tau).
\end{aligned}
\label{eq:Jsgrad}
\end{equation}}\ignorespaces
The field $G_s$ thus acts as a target for the call's velocity at its shadow input. It involves three quantities that we cannot evaluate: the reward gradient $\nabla R$ in $G_s^{\mathrm R}$, as well as the velocities $v_{\hat q_s}$ and $v_q$ in $G_s^{\mathrm D}$. Both velocities belong to distributions that we can sample. We therefore replace them by DMD's fake score $F$, to which we add a per-call condition $c$. The condition is a stop for that call's outputs or the empty condition $\varnothing$ for the final samples. Since its flow-matching loss splits into one loss per condition, training on the samples of each condition gives $F(y,t)=F(y,t;\varnothing)\approx v_q$ and $\hat F_s(y,t)=F(y,t;s)\approx v_{\hat q_s}$.

\vspace{3pt}
\noindent\textbf{Shadow credit.} For the reward gradient, we turn to the reward values of the final samples. Given the shadow input, the call's output $\hat x_s$ is fixed, so its single reward value carries no information about $\nabla R(\hat x_s)$. The final sample $x$, in contrast, varies over the shadow posterior, so its reward carries information about the average gradient $\E[\nabla R(x)\mid y_s]$, which we use in place of $\nabla R(\hat x_s)$. This average still involves $\nabla R$, but \Cref{eq:Jsgrad} needs only the conditional mean of $G_s$ given $y_s$, as $\partial_\theta v_\theta(y_s,s)$ is a function of $y_s$. Hence $G_s^{\mathrm R}$, now $-s\,\E[\nabla R(x)\mid y_s]$, can be replaced by any term with this conditional mean without changing $\nabla_\theta J_s$.

\begin{lemma}[Shadow credit]
\label{lem:credit}
Let $r=u-v_q(y_s,s)$ and $V_s=\Cov(u\mid y_s)=\E[rr^\top\mid y_s]$. Then $V_s^{-1}\E[R\,r\mid y_s]=-s\,\E[\nabla R(x)\mid y_s]$ if $R$ is affine, or if the shadow posterior is Gaussian and $R$ is smooth with integrable gradient.
\end{lemma}
\vspace{3pt}
\begin{proof}
Since $v_q(y_s,s)=\E[u\mid y_s]$ and $x=y_s-s\,u$, we have $x-\E[x\mid y_s]=-s\,r$. Hence $\E[R\,r\mid y_s]=-\E[R(x)\,(x-\E[x\mid y_s])\mid y_s]/s$. This expectation equals $\Cov(x\mid y_s)\,\E[\nabla R(x)\mid y_s]$, by linearity for affine $R$ and by Stein's lemma for a Gaussian posterior. As $\Cov(x\mid y_s)=s^2V_s$, inverting the nonsingular $V_s$ gives the claim.
\end{proof}
\vspace{3pt}
By \Cref{lem:credit}, $R\,V_s^{-1}r$ is such a term, and it uses only reward values. Since $\E[r\mid y_s]=0$, its conditional mean equals $V_s^{-1}\Cov(R,u\mid y_s)$, the covariance term of \Cref{eq:consistency} with $w=R$, rescaled by $V_s^{-1}$. For the same reason, a baseline $b$ that does not depend on the sample leaves this mean unchanged. With the advantage $A=R-b$ and $F$ in place of $v_q$, the shadow credit enters the target at stop $s$ as:
\begin{equation}
G_s^{\mathrm R}=A\,V_s^{-1}\big(u-F(y_s,s)\big).
\label{eq:credit}
\end{equation}

\noindent\textbf{Remarks.} Since the shadow posterior is non-degenerate at every stop, the shadow credit reaches every call. At the final call, it replaces the pathwise term of \Cref{eq:pg} and needs only reward values. Its reference is $v_q$ rather than $v_\theta$, so by \Cref{eq:consistency} it adds no reward-free shift and keeps the samples sharp. Unlike in \Cref{eq:rwdm}, it also does not vanish at the teacher, where both divergences do. The reward thus provides a direction independent of the anchor's force.

\noindent\textbf{Adaptive weighting.} The shadow credit of \Cref{eq:credit} still contains the unknown precision $V_s^{-1}$, which varies with the stop, the prompt, and the sample, so neither a constant nor a per-stop statistic can replace it. In \Cref{lem:credit}, however, it serves only to normalize the second moment $V_s=\E[rr^\top\mid y_s]$ of the residual $r$. This suggests replacing it by a per-sample normalizer $1/Q$ with $Q={\sg\big[\|u-F(y_s,s)\|^2/d\big]}$, where $d$ is the number of latent coordinates and $\sg$ the stop-gradient. When $F=v_q$ and $r$ is spherically symmetric given $y_s$, this normalizer has the same effect as $V_s^{-1}$: it normalizes the second moment, $\E[rr^\top/Q\mid y_s]=I$, as $rr^\top/Q$ depends only on the uniform direction of $r$, and leaves the first moment zero, $\E[r/Q\mid y_s]=0$, as $r/Q$ is odd in $r$. The credit thus keeps its conditional mean for quadratic rewards. Among denominators that are functions of $\|r\|$, only the mean square does so for every law of $\|r\|$.

The divergence terms still contain the time weight $\kappa(\tau)$, which ranges over several orders of magnitude on the support of $\nu$. Following DMD~\cite{yin2024onestep}, we instead normalize each sample's force by its own scale, replacing $\kappa(\tau)$ by $1/Z$ with $Z=\sg\big[\|v_T(\hat y,\tau)-(\xi-\hat x_s)\|_1/d\big]$.

With $1/Q$ and $1/Z$, the target at stop $s$ becomes:
\begingroup
\setlength{\abovedisplayskip}{6pt}
\setlength{\abovedisplayshortskip}{6pt}
\begin{equation}
\begin{aligned}
G_s={}&\frac{A}{Q}\,\big(u-F(y_s,s)\big)\\
&+\frac{s}{Z}\Big[\alpha\big(v_T-\hat F_s\big)+\gamma\big(F-\hat F_s\big)\Big](\hat y,\tau).
\end{aligned}
\label{eq:target}
\end{equation}
\endgroup

\noindent\textbf{Surrogate loss.} We minimize the regression loss at stop $s$:
\begin{equation}
\mathcal L_s(\theta)=\frac12\,\E\Big[\big\|v_\theta(y_s,s)-\sg\big[v_\theta(y_s,s)+G_s\big]\big\|^2\Big].
\label{eq:loss}
\end{equation}
Since the stop-gradient copy carries no gradient, the negative gradient of $\mathcal L_s$ is $\E[G_s\cdot\partial_\theta v_\theta(y_s,s)]$, which is \Cref{eq:Jsgrad} with $G_s$ given by \Cref{eq:target}.

\begin{algorithm}[t]
\caption{Shadow Posterior Matching.}
\label{alg:spm}
\linespread{1.1}\selectfont
\algrenewcommand\algorithmicindent{1em}
\begin{algorithmic}[1]
\Require DMD-distilled generator $v_\theta$ and fake score $F$, teacher $v_T$, deployed stops $s_1,\dots,s_K$, $\alpha$, $\gamma$
\State Add the condition $c$ to $F$, with all its vectors set to zero
\While{not converged}
  \State Sample prompts and $N$ final samples $x$ per prompt with the $K$-step sampler, without gradients \Comment{rollout}
  \State Score $x$ with the black-box reward and set $A$ to the clipped group-relative advantages
  \State Draw a deployed stop $s$ uniformly \Comment{fake score}
  \State $\epsilon\sim\N(0,I)$, $y\gets(1-s)x+s\epsilon$, $\hat x\gets y-s\,v_\theta(y,s)$
  \State Update $F$ by flow matching on $(x,\varnothing)$ and $(\hat x,s)$
  \State Draw a deployed stop $s$ uniformly \Comment{generator}
  \State $\epsilon\sim\N(0,I)$, $y\gets(1-s)x+s\epsilon$, $u\gets\epsilon-x$
  \State $\hat x\gets y-s\,v_\theta(y,s)$, $\tau\sim\nu$, $\xi\sim\N(0,I)$,
  \Statex \hspace{\algorithmicindent}$\hat y\gets(1-\tau)\,\sg[\hat x]+\tau\xi$
  \State $G\gets$ \Cref{eq:target} at $(y,s)$ and $(\hat y,\tau)$
  \State Update $\theta$ on \Cref{eq:loss}
\EndWhile
\end{algorithmic}
\end{algorithm}

\subsection{Training}
\label{sec:train}
As illustrated in \Cref{fig:method} and summarized in \Cref{alg:spm}, each training iteration consists of a rollout of the deployed sampler, a fake-score update, and a generator update.

\vspace{3pt}
\noindent\textbf{Rollout.} For every prompt, the $K$-step sampler of \Cref{eq:sampler} generates a group of final samples with the current generator, without gradients. We score each final sample with the black-box reward and normalize the scores to obtain the clipped group-relative advantages~\cite{shao2024deepseekmath}.

\vspace{3pt}
\noindent\textbf{Fake score.} The fake score is then updated by flow matching on the same rollout without using the rewards. A single update sums the losses on the final samples under $\varnothing$ and on the call's outputs at one deployed stop $s$ under $s$. The condition of each stop is a learnable vector added to the time embedding and initialized to zero. At initialization, $\hat F_s=F$, so endpoint matching exerts no force.

\vspace{3pt}
\noindent\textbf{Generator.} We train the generator only at the deployed stops, since it is called nowhere else. Each update draws one of these stops uniformly. At this stop, the generator descends \Cref{eq:loss}, with the target \Cref{eq:target} computed from the updated fake score and the teacher. Only its velocity at the shadow input carries a gradient.

\section{Experiments}
\label{sec:exp}

\begin{table*}[t]
  \captionsetup{skip=4pt}
  \centering
  \caption{\textbf{Quantitative results.} We report training rewards on test prompts, and image-quality metrics and preference scores on DrawBench prompts. Higher is better for all metrics. The best 4-NFE results within each task are shown in \textbf{bold} and second-best results are \underline{underlined}.}
  \label{tab:main_results}
  \small
  \setlength{\tabcolsep}{4.5pt}
  \renewcommand{\arraystretch}{1.1}
  \begin{tabular}{@{}lccccccccc@{}}
    \toprule
    \multirow{2}{*}[-0.65ex]{\textbf{Method}} & \multirow{2}{*}[-0.65ex]{\textbf{NFE}} & \multicolumn{3}{c}{\textbf{Training Reward}} & \multicolumn{2}{c}{\textbf{Image Quality}} & \multicolumn{3}{c}{\textbf{Preference Score}} \\
    \cmidrule(lr){3-5}
    \cmidrule(lr){6-7}
    \cmidrule(lr){8-10}
    & & \textbf{GenEval} & \textbf{OCR} & \textbf{PickScore} & \textbf{Aesthetic} & \textbf{DeQA} & \textbf{ImageReward} & \textbf{PickScore} & \textbf{HPSv2.1} \\
    \midrule
    SD3.5-M~\cite{esser2024scaling} & 80 & 0.64 & 0.56 & 21.79 & 5.41 & 4.08 & 0.82 & 22.40 & 0.28 \\
    \midrule
    DMD2~\cite{yin2024improved} & 4 & 0.69 & 0.59 & 21.78 & 5.50 & 4.16 & 1.09 & 22.48 & 0.30 \\
    \midrule
    \multicolumn{10}{c}{\textit{Compositional Image Generation}} \\
    \midrule
    Flow-GRPO~\cite{liu2025flowgrpo} & 80 & 0.95 & -- & -- & 5.25 & 4.01 & 1.03 & 22.37 & 0.27 \\
    DiffusionNFT~\cite{zheng2026diffusionnft} & 40 & 0.95 & -- & -- & 4.98 & 4.10 & 0.30 & 21.59 & 0.24 \\
    RAM~\cite{bergmeister2026reinforce} & 80 & 0.97 & -- & -- & 5.38 & 4.09 & 1.19 & 22.52 & 0.29 \\
    \midrule
    RTDMD~\cite{huang2026reinforcing} & 4 & 0.77 & -- & -- & \textbf{5.52} & \textbf{4.15} & \underline{0.97} & \underline{22.30} & \underline{0.28} \\
    TDM-R1~\cite{luo2026tdmr1} & 4 & \underline{0.86} & -- & -- & 5.28 & 3.52 & 0.86 & 21.90 & 0.27 \\
    \method (Ours) & 4 & \textbf{0.94} & -- & -- & \underline{5.30} & \underline{4.09} & \textbf{1.07} & \textbf{22.49} & \textbf{0.29} \\
    \midrule
    \multicolumn{10}{c}{\textit{Visual Text Rendering}} \\
    \midrule
    Flow-GRPO~\cite{liu2025flowgrpo} & 80 & -- & 0.92 & -- & 5.32 & 4.06 & 0.95 & 22.44 & 0.28 \\
    DiffusionNFT~\cite{zheng2026diffusionnft} & 40 & -- & 0.96 & -- & 4.87 & 3.01 & -0.97 & 20.26 & 0.18 \\
    RAM~\cite{bergmeister2026reinforce} & 80 & -- & 0.97 & -- & 5.23 & 3.90 & 0.44 & 21.83 & 0.26 \\
    \midrule
    RTDMD~\cite{huang2026reinforcing} & 4 & -- & 0.92 & -- & 5.37 & 3.54 & 0.74 & 21.65 & 0.25 \\
    TDM-R1~\cite{luo2026tdmr1} & 4 & -- & \underline{0.95} & -- & \underline{5.50} & \underline{4.10} & \underline{1.10} & \underline{22.43} & \underline{0.30} \\
    \method (Ours) & 4 & -- & \textbf{0.96} & -- & \textbf{5.59} & \textbf{4.21} & \textbf{1.14} & \textbf{22.63} & \textbf{0.31} \\
    \midrule
    \multicolumn{10}{c}{\textit{Human Preference Alignment}} \\
    \midrule
    Flow-GRPO~\cite{liu2025flowgrpo} & 80 & -- & -- & 23.31 & 5.92 & 4.22 & 1.28 & 23.53 & 0.32 \\
    DiffusionNFT~\cite{zheng2026diffusionnft} & 40 & -- & -- & 23.29 & 6.16 & 4.13 & 1.23 & 23.65 & 0.31 \\
    RAM~\cite{bergmeister2026reinforce} & 80 & -- & -- & 23.67 & 6.11 & 4.17 & 1.36 & 23.95 & 0.32 \\
    \midrule
    RTDMD~\cite{huang2026reinforcing} & 4 & -- & -- & \underline{23.36} & \underline{6.07} & \textbf{4.27} & 1.25 & \underline{23.64} & \textbf{0.32} \\
    TDM-R1~\cite{luo2026tdmr1} & 4 & -- & -- & 23.18 & 5.76 & 4.21 & \textbf{1.34} & 23.43 & \underline{0.32} \\
    \method (Ours) & 4 & -- & -- & \textbf{23.84} & \textbf{6.34} & \underline{4.21} & \underline{1.32} & \textbf{24.04} & 0.31 \\
    \bottomrule
  \end{tabular}
  \vspace{-4pt}
\end{table*}

\subsection{Experimental Setup}

\noindent\textbf{Tasks and evaluation.} We compare \method with the state-of-the-art few-step RL methods RTDMD~\cite{huang2026reinforcing} and TDM-R1~\cite{luo2026tdmr1} on SD3.5-Medium (SD3.5-M)~\cite{esser2024scaling}. We study three tasks: compositional image generation with GenEval~\cite{ghosh2023geneval}, visual text rendering with the OCR reward~\cite{liu2025flowgrpo}, and human preference alignment with PickScore~\cite{kirstain2023pickapic}. All few-step methods start from the same 4-step generator and fake score, distilled with DMD2~\cite{yin2024improved} on the training prompts of all three tasks. For each task, we fine-tune this generator on the task's training prompts and report the reward on its test prompts. To measure reward hacking, we further evaluate on the DrawBench prompts~\cite{saharia2022photorealistic} with two image-quality metrics, Aesthetic Score~\cite{schuhmann2022laion5b} and DeQA~\cite{you2025teaching}, and three preference scores, ImageReward~\cite{xu2023imagereward}, PickScore~\cite{kirstain2023pickapic}, and HPSv2.1~\cite{wu2023human}. For reference, we also list the multi-step teacher and the multi-step RL methods Flow-GRPO~\cite{liu2025flowgrpo}, DiffusionNFT~\cite{zheng2026diffusionnft}, and RAM~\cite{bergmeister2026reinforce}, whose number of function evaluations (NFE) is 40 or 80.

\vspace{3pt}
\noindent\textbf{Implementation details.} We implement the baselines in a unified codebase, following their official repositories. The generator is deployed at the stops $1$, $0.9$, $0.75$, and $0.5$, obtained by applying a time shift of $3$ to a uniform grid. For \method, we set $\alpha=\gamma=0.1$. All few-step methods train for 1,000 iterations at 512 resolution on 8 NVIDIA H100 GPUs. Each iteration draws 24 samples per prompt from the 4-step sampler for a batch of 48 prompts. It normalizes the 24 rewards of every prompt into group-relative advantages clipped to $[-5,5]$ and then updates the fake score and the generator once. The generator and the fake score are fine-tuned with LoRA (rank 32, alpha 64) and optimized by AdamW ($\beta_1=0.9$, $\beta_2=0.999$) in mixed precision, with learning rates of $3\times10^{-6}$ for the generator and $3\times10^{-5}$ for the fake score. For evaluation, we use an exponential moving average (EMA) of the generator with decay $0.99$.

\subsection{Main Results}

\noindent\textbf{Quantitative results.} \Cref{tab:main_results} shows that \method achieves the highest training reward among the 4-step methods on all three tasks: 0.94 on GenEval, 0.96 on OCR, and 23.84 on PickScore, against 0.86, 0.95, and 23.36 for the strongest baseline. In the reward curves of \Cref{fig:teaser}, \method pulls ahead early and stays ahead throughout training on GenEval and OCR. More importantly, this higher reward comes with less reward hacking. On DrawBench, \method scores best among the 4-step methods on 10 of the 15 task--metric pairs. On OCR, it even improves all five metrics over DMD2, so its gain in reward costs no image quality. With 4 NFEs instead of 40 or 80, \method clearly exceeds the multi-step teacher on all three rewards and approaches or surpasses the state-of-the-art multi-step RL methods.

These gaps follow from how each method credits the reward (\Cref{sec:background}). On the black-box GenEval and OCR rewards, RTDMD credits only the renoising kernels, through high-variance likelihood ratios, and leaves the deterministic final call to the teacher anchor alone (\Cref{eq:pg}), which limits its reward. TDM-R1 reaches the final call only through an online proxy of the reward, which the generator exploits where the fit is poor. Rule-based rewards such as GenEval and OCR are hard to fit, so TDM-R1 shows noticeable reward hacking on them. The shadow credit of \method instead uses measured reward values and reaches every call through the non-degenerate shadow posterior. The final call thus receives credit, which raises the reward on both tasks. Without a fitted proxy, the generator also has no fitting error to exploit, which reduces reward hacking.

\begin{figure*}[t]
\captionsetup{skip=4pt}
\centering
\includegraphics[width=\textwidth]{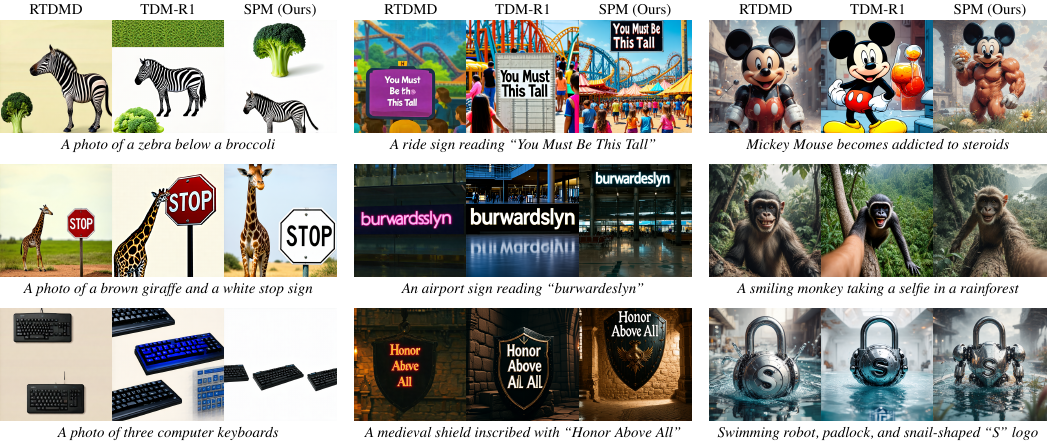}
\caption{\textbf{Qualitative comparison.} Samples are generated from GenEval (left), OCR (center), and PickScore (right) test prompts.}
\label{fig:qualitative}
\vspace{-4pt}
\end{figure*}

\vspace{3pt}
\noindent\textbf{Qualitative results.} \Cref{fig:qualitative} compares samples of few-step RL methods. \method composes objects with the correct spatial relations, colors, and counts (GenEval), renders the specified text accurately (OCR), and depicts the key concepts of the prompts (PickScore). In contrast, the baselines often fall short in these respects and show signs of reward hacking: TDM-R1 produces over-sharpened edges and fragmented objects on GenEval, and RTDMD produces blurred, low-detail scenes on OCR.

\begin{figure*}[t]
\captionsetup{skip=4pt}
\centering
\includegraphics{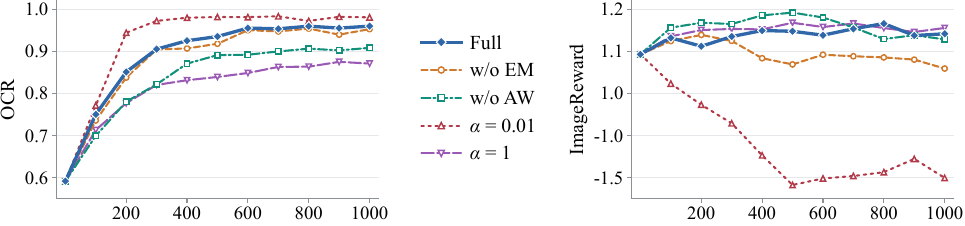}
\caption{\textbf{Ablation results on visual text rendering.} \textbf{Left:} OCR reward on test prompts. \textbf{Right:} ImageReward on DrawBench prompts. EM and AW denote endpoint matching and adaptive weighting, respectively. $\alpha$ is the tilting temperature, with a default value of 0.1.}
\label{fig:ablation}
\vspace{-4pt}
\end{figure*}

\subsection{Ablation Study}
\label{sec:ablation}
We ablate three aspects of \method on visual text rendering with the OCR reward and plot the results in \Cref{fig:ablation}.

\vspace{3pt}
\noindent\textbf{Effect of endpoint matching.} Removing EM from \Cref{eq:Js} lowers the OCR reward to 0.95 and ImageReward to 1.06. The two runs coincide for the first 300 training steps. The outputs of the earlier calls then drift from the final distribution, so the later calls receive deployed inputs that differ from their shadow inputs (\Cref{lem:induction}). Since the teacher anchor and the shadow credit act only on the shadow inputs, the sampler loses image quality without gaining reward. EM keeps the two inputs aligned, so the teacher anchor preserves the image quality of the deployed sampler.

\noindent\textbf{Effect of adaptive weighting.} We replace the adaptive weight $1/Q$ of the shadow credit in \Cref{eq:target} by the precision $V_s^{-1}$ of \Cref{lem:credit}. Since the true $V_s$ is unknown, we set $V_s^{-1}$ to $(1-s)^2+s^2$, the value for a standard Gaussian final distribution. The OCR reward then drops to 0.91, while ImageReward stays at 1.13. The Gaussian value depends only on the stop, but the spread of the shadow posterior also varies with the prompt and the sample. The credit is therefore too strong for some samples and too weak for others. The weight $1/Q$ instead adapts to each sample through its residual. For quadratic rewards and spherically symmetric residuals, it keeps the conditional mean of the credit whatever the spread. This result shows that the per-sample scale improves the shadow credit itself, raising the reward at no cost in image quality.

\vspace{3pt}
\noindent\textbf{Effect of tilting temperature.} We vary $\alpha\in\{0.01,0.1,1\}$. At $\alpha=0.01$, the OCR reward rises to 0.98, but ImageReward collapses to $-1.50$. At $\alpha=1$, ImageReward rises to 1.16, but the OCR reward reaches only 0.87. Since $\alpha$ weights the teacher anchor in \Cref{eq:Js}, a small $\alpha$ lets the generator move far from the teacher, into regions where the reward can be exploited. A large one keeps it close to the teacher and leaves little room to raise the reward. The tilting temperature therefore sets the trade-off between reward maximization and image quality.

\section{Conclusion}
\label{sec:concl}

We study RL with black-box rewards for few-step diffusion models and identify two challenges: the velocity mismatch and the external anchor. Existing few-step RL methods meet the latter with a marginal KL to the teacher but pay for the former in their reward credit. \method instead credits every call through the shadow posterior of generated samples. From per-call KL-regularized reward objectives, we derive a regression loss with three terms: shadow credit, teacher anchor, and endpoint matching. On few-step generators distilled by distribution matching, \method achieves higher rewards with less reward hacking than prior few-step RL methods. Since the shadow credit needs only generated samples and a score of their distribution, we expect it to extend to other few-step generators, such as consistency models, and leave this extension to future work.

\par
{
    \small
    \bibliographystyle{ieeenat_fullname}
    \bibliography{main}

\begin{thebibliography}{44}
\providecommand{\natexlab}[1]{#1}
\providecommand{\url}[1]{\texttt{#1}}
\expandafter\ifx\csname urlstyle\endcsname\relax
  \providecommand{\doi}[1]{doi: #1}\else
  \providecommand{\doi}{doi: \begingroup \urlstyle{rm}\Url}\fi

\bibitem[Bai et~al.(2026)Bai, Zhou, Shao, Zhong, Yang, Chen, Cheng, and
  Xie]{bai2026optimizing}
Lichen Bai, Zikai Zhou, Shitong Shao, Wenliang Zhong, Shuo Yang, Shuo Chen,
  Bojun Cheng, and Zeke Xie.
\newblock {Optimizing Few-Step Generation with Adaptive Matching Distillation}.
\newblock In \emph{International Conference on Machine Learning}, 2026.

\bibitem[Bergmeister et~al.(2026)Bergmeister, Jegelka, N{\"u}sken,
  Domingo-Enrich, and Pidstrigach]{bergmeister2026reinforce}
Andreas Bergmeister, Stefanie Jegelka, Nikolas N{\"u}sken, Carles
  Domingo-Enrich, and Jakiw Pidstrigach.
\newblock {Reinforce Adjoint Matching: Scaling RL Post-Training of Diffusion
  and Flow-Matching Models}.
\newblock In \emph{Advances in Neural Information Processing Systems}, 2026.

\bibitem[Black et~al.(2024)Black, Janner, Du, Kostrikov, and
  Levine]{black2024training}
Kevin Black, Michael Janner, Yilun Du, Ilya Kostrikov, and Sergey Levine.
\newblock {Training Diffusion Models with Reinforcement Learning}.
\newblock In \emph{International Conference on Learning Representations}, 2024.

\bibitem[Clark et~al.(2024)Clark, Vicol, Swersky, and Fleet]{clark2024directly}
Kevin Clark, Paul Vicol, Kevin Swersky, and David~J. Fleet.
\newblock {Directly Fine-Tuning Diffusion Models on Differentiable Rewards}.
\newblock In \emph{International Conference on Learning Representations}, 2024.

\bibitem[Domingo-Enrich et~al.(2025)Domingo-Enrich, Drozdzal, Karrer, and
  Chen]{domingoenrich2025adjoint}
Carles Domingo-Enrich, Michal Drozdzal, Brian Karrer, and Ricky T.~Q. Chen.
\newblock {Adjoint Matching: Fine-Tuning Flow and Diffusion Generative Models
  with Memoryless Stochastic Optimal Control}.
\newblock In \emph{International Conference on Learning Representations}, 2025.

\bibitem[Dong et~al.(2026)Dong, Guo, Bai, Zheng, Luo, and Zou]{dong2026guiding}
Linwei Dong, Ruoyu Guo, Ge Bai, Quan Zheng, Yawei Luo, and Changqing Zou.
\newblock {Guiding Distribution Matching Distillation with Gradient-Based
  Reinforcement Learning}.
\newblock In \emph{Advances in Neural Information Processing Systems}, 2026.

\bibitem[Esser et~al.(2024)Esser, Kulal, Blattmann, Entezari, M{\"u}ller,
  Saini, Levi, Lorenz, Sauer, Boesel, Podell, Dockhorn, English, and
  Rombach]{esser2024scaling}
Patrick Esser, Sumith Kulal, Andreas Blattmann, Rahim Entezari, Jonas
  M{\"u}ller, Harry Saini, Yam Levi, Dominik Lorenz, Axel Sauer, Frederic
  Boesel, Dustin Podell, Tim Dockhorn, Zion English, and Robin Rombach.
\newblock {Scaling Rectified Flow Transformers for High-Resolution Image
  Synthesis}.
\newblock In \emph{International Conference on Machine Learning}, 2024.

\bibitem[Fan et~al.(2025)Fan, Shen, Cheng, Chen, Liang, and Liu]{fan2025online}
Jiajun Fan, Shuaike Shen, Chaoran Cheng, Yuxin Chen, Chumeng Liang, and Ge Liu.
\newblock {Online Reward-Weighted Fine-Tuning of Flow Matching with Wasserstein
  Regularization}.
\newblock In \emph{International Conference on Learning Representations}, 2025.

\bibitem[Fan et~al.(2023)Fan, Watkins, Du, Liu, Ryu, Boutilier, Abbeel,
  Ghavamzadeh, Lee, and Lee]{fan2023dpok}
Ying Fan, Olivia Watkins, Yuqing Du, Hao Liu, Moonkyung Ryu, Craig Boutilier,
  Pieter Abbeel, Mohammad Ghavamzadeh, Kangwook Lee, and Kimin Lee.
\newblock {DPOK: Reinforcement Learning for Fine-Tuning Text-to-Image Diffusion
  Models}.
\newblock In \emph{Advances in Neural Information Processing Systems}, 2023.

\bibitem[Ghosh et~al.(2023)Ghosh, Hajishirzi, and Schmidt]{ghosh2023geneval}
Dhruba Ghosh, Hannaneh Hajishirzi, and Ludwig Schmidt.
\newblock {GenEval: An Object-Focused Framework for Evaluating Text-to-Image
  Alignment}.
\newblock In \emph{Advances in Neural Information Processing Systems}, 2023.

\bibitem[Huang et~al.(2026)Huang, Zhou, Wang, Zhang, Zhang, and
  Pang]{huang2026reinforcing}
Yushi Huang, Xiangxin Zhou, Ruoyu Wang, Chi Zhang, Jun Zhang, and Tianyu Pang.
\newblock {Reinforcing Few-Step Generators via Reward-Tilted Distribution
  Matching}.
\newblock \emph{arXiv preprint arXiv:2605.26108}, 2026.

\bibitem[Jia et~al.(2025)Jia, Nan, Zhao, and Liu]{jia2025reward}
Zhiwei Jia, Yuesong Nan, Huixi Zhao, and Gengdai Liu.
\newblock {Reward Fine-Tuning Two-Step Diffusion Models via Learning
  Differentiable Latent-Space Surrogate Reward}.
\newblock In \emph{IEEE/CVF Conference on Computer Vision and Pattern
  Recognition}, 2025.

\bibitem[Jiang et~al.(2026)Jiang, Liu, Wang, Wu, Li, Li, Jin, Lu, Li, Wang,
  et~al.]{jiang2026distribution}
Dengyang Jiang, Dongyang Liu, Zanyi Wang, Qilong Wu, Liuzhuozheng Li,
  Hengzhuang Li, Xin Jin, Changsheng Lu, Zhen Li, Mengmeng Wang, et~al.
\newblock {Distribution Matching Distillation Meets Reinforcement Learning}.
\newblock In \emph{European Conference on Computer Vision}, 2026.

\bibitem[Kirstain et~al.(2023)Kirstain, Polyak, Singer, Matiana, Penna, and
  Levy]{kirstain2023pickapic}
Yuval Kirstain, Adam Polyak, Uriel Singer, Shahbuland Matiana, Joe Penna, and
  Omer Levy.
\newblock {Pick-a-Pic: An Open Dataset of User Preferences for Text-to-Image
  Generation}.
\newblock In \emph{Advances in Neural Information Processing Systems}, 2023.

\bibitem[Lee et~al.(2023)Lee, Liu, Ryu, Watkins, Du, Boutilier, Abbeel,
  Ghavamzadeh, and Gu]{lee2023aligning}
Kimin Lee, Hao Liu, Moonkyung Ryu, Olivia Watkins, Yuqing Du, Craig Boutilier,
  Pieter Abbeel, Mohammad Ghavamzadeh, and Shixiang~Shane Gu.
\newblock {Aligning Text-to-Image Models Using Human Feedback}.
\newblock \emph{arXiv preprint arXiv:2302.12192}, 2023.

\bibitem[Liu et~al.(2025)Liu, Liu, Liang, Li, Liu, Wang, Wan, Zhang, and
  Ouyang]{liu2025flowgrpo}
Jie Liu, Gongye Liu, Jiajun Liang, Yangguang Li, Jiaheng Liu, Xintao Wang,
  Pengfei Wan, Di Zhang, and Wanli Ouyang.
\newblock {Flow-GRPO: Training Flow Matching Models via Online RL}.
\newblock In \emph{Advances in Neural Information Processing Systems}, 2025.

\bibitem[Lu and Song(2025)]{lu2025simplifying}
Cheng Lu and Yang Song.
\newblock {Simplifying, Stabilizing and Scaling Continuous-Time Consistency
  Models}.
\newblock In \emph{International Conference on Learning Representations}, 2025.

\bibitem[Lu et~al.(2026)Lu, Zeng, Li, Ouyang, Wang, Cheng, Zhu, Cao, Zhang,
  Zhu, et~al.]{lu2025reward}
Yunhong Lu, Yanhong Zeng, Haobo Li, Hao Ouyang, Qiuyu Wang, Ka~Leong Cheng,
  Jiapeng Zhu, Hengyuan Cao, Zhipeng Zhang, Xing Zhu, et~al.
\newblock {Reward Forcing: Efficient Streaming Video Generation with Rewarded
  Distribution Matching Distillation}.
\newblock In \emph{IEEE/CVF Conference on Computer Vision and Pattern
  Recognition}, 2026.

\bibitem[Luo et~al.(2023{\natexlab{a}})Luo, Tan, Huang, Li, and
  Zhao]{luo2023latent}
Simian Luo, Yiqin Tan, Longbo Huang, Jian Li, and Hang Zhao.
\newblock {Latent Consistency Models: Synthesizing High-Resolution Images with
  Few-Step Inference}.
\newblock \emph{arXiv preprint arXiv:2310.04378}, 2023{\natexlab{a}}.

\bibitem[Luo(2024)]{luo2024diffinstruct}
Weijian Luo.
\newblock {Diff-Instruct++: Training One-Step Text-to-Image Generator Model to
  Align with Human Preferences}.
\newblock \emph{Transactions on Machine Learning Research}, 2024.

\bibitem[Luo et~al.(2023{\natexlab{b}})Luo, Hu, Zhang, Sun, Li, and
  Zhang]{luo2023diffinstruct}
Weijian Luo, Tianyang Hu, Shifeng Zhang, Jiacheng Sun, Zhenguo Li, and Zhihua
  Zhang.
\newblock {Diff-Instruct: A Universal Approach for Transferring Knowledge from
  Pre-Trained Diffusion Models}.
\newblock In \emph{Advances in Neural Information Processing Systems},
  2023{\natexlab{b}}.

\bibitem[Luo et~al.(2025)Luo, Hu, Sun, Cai, and Tang]{luo2025learning}
Yihong Luo, Tianyang Hu, Jiacheng Sun, Yujun Cai, and Jing Tang.
\newblock {Learning Few-Step Diffusion Models by Trajectory Distribution
  Matching}.
\newblock In \emph{IEEE/CVF International Conference on Computer Vision}, 2025.

\bibitem[Luo et~al.(2026)Luo, Hu, Luo, and Tang]{luo2026tdmr1}
Yihong Luo, Tianyang Hu, Weijian Luo, and Jing Tang.
\newblock {TDM-R1: Reinforcing Few-Step Diffusion Models with
  Non-Differentiable Reward}.
\newblock In \emph{International Conference on Machine Learning}, 2026.

\bibitem[Oertell et~al.(2024)Oertell, Chang, Zhang, Brantley, and
  Sun]{oertell2024rl}
Owen Oertell, Jonathan~Daniel Chang, Yiyi Zhang, Kiant{\'e} Brantley, and Wen
  Sun.
\newblock {RL for Consistency Models: Reward Guided Text-to-Image Generation
  with Fast Inference}.
\newblock \emph{Reinforcement Learning Journal}, 2024.

\bibitem[Prabhudesai et~al.(2023)Prabhudesai, Goyal, Pathak, and
  Fragkiadaki]{prabhudesai2023aligning}
Mihir Prabhudesai, Anirudh Goyal, Deepak Pathak, and Katerina Fragkiadaki.
\newblock {Aligning Text-to-Image Diffusion Models with Reward
  Backpropagation}.
\newblock \emph{arXiv preprint arXiv:2310.03739}, 2023.

\bibitem[Ren et~al.(2024)Ren, Xia, Lu, Zhang, Wu, Xie, Wang, and
  Xiao]{ren2024hypersd}
Yuxi Ren, Xin Xia, Yanzuo Lu, Jiacheng Zhang, Jie Wu, Pan Xie, Xing Wang, and
  Xuefeng Xiao.
\newblock {Hyper-SD: Trajectory Segmented Consistency Model for Efficient Image
  Synthesis}.
\newblock In \emph{Advances in Neural Information Processing Systems}, 2024.

\bibitem[Saharia et~al.(2022)Saharia, Chan, Saxena, Li, Whang, Denton,
  Ghasemipour, Lopes, Ayan, Salimans, et~al.]{saharia2022photorealistic}
Chitwan Saharia, William Chan, Saurabh Saxena, Lala Li, Jay Whang, Emily~L.
  Denton, Kamyar Ghasemipour, Raphael~Gontijo Lopes, Burcu~Karagol Ayan, Tim
  Salimans, et~al.
\newblock {Photorealistic Text-to-Image Diffusion Models with Deep Language
  Understanding}.
\newblock In \emph{Advances in Neural Information Processing Systems}, 2022.

\bibitem[Salimans and Ho(2022)]{salimans2022progressive}
Tim Salimans and Jonathan Ho.
\newblock {Progressive Distillation for Fast Sampling of Diffusion Models}.
\newblock In \emph{International Conference on Learning Representations}, 2022.

\bibitem[Sauer et~al.(2024{\natexlab{a}})Sauer, Boesel, Dockhorn, Blattmann,
  Esser, and Rombach]{sauer2024fast}
Axel Sauer, Frederic Boesel, Tim Dockhorn, Andreas Blattmann, Patrick Esser,
  and Robin Rombach.
\newblock {Fast High-Resolution Image Synthesis with Latent Adversarial
  Diffusion Distillation}.
\newblock In \emph{ACM SIGGRAPH Asia Conference}, 2024{\natexlab{a}}.

\bibitem[Sauer et~al.(2024{\natexlab{b}})Sauer, Lorenz, Blattmann, and
  Rombach]{sauer2024adversarial}
Axel Sauer, Dominik Lorenz, Andreas Blattmann, and Robin Rombach.
\newblock {Adversarial Diffusion Distillation}.
\newblock In \emph{European Conference on Computer Vision}, 2024{\natexlab{b}}.

\bibitem[Schuhmann et~al.(2022)Schuhmann, Beaumont, Vencu, Gordon, Wightman,
  Cherti, Coombes, Katta, Mullis, Wortsman, et~al.]{schuhmann2022laion5b}
Christoph Schuhmann, Romain Beaumont, Richard Vencu, Cade Gordon, Ross
  Wightman, Mehdi Cherti, Theo Coombes, Aarush Katta, Clayton Mullis, Mitchell
  Wortsman, et~al.
\newblock {LAION-5B: An Open Large-Scale Dataset for Training Next Generation
  Image-Text Models}.
\newblock In \emph{Advances in Neural Information Processing Systems}, 2022.

\bibitem[Shao et~al.(2024)Shao, Wang, Zhu, Xu, Song, Bi, Zhang, Zhang, Li, Wu,
  et~al.]{shao2024deepseekmath}
Zhihong Shao, Peiyi Wang, Qihao Zhu, Runxin Xu, Junxiao Song, Xiao Bi, Haowei
  Zhang, Mingchuan Zhang, Y.~K. Li, Y. Wu, et~al.
\newblock {DeepSeekMath: Pushing the Limits of Mathematical Reasoning in Open
  Language Models}.
\newblock \emph{arXiv preprint arXiv:2402.03300}, 2024.

\bibitem[Song et~al.(2023)Song, Dhariwal, Chen, and
  Sutskever]{song2023consistency}
Yang Song, Prafulla Dhariwal, Mark Chen, and Ilya Sutskever.
\newblock {Consistency Models}.
\newblock In \emph{International Conference on Machine Learning}, 2023.

\bibitem[Wallace et~al.(2024)Wallace, Dang, Rafailov, Zhou, Lou, Purushwalkam,
  Ermon, Xiong, Joty, and Naik]{wallace2024diffusion}
Bram Wallace, Meihua Dang, Rafael Rafailov, Linqi Zhou, Aaron Lou, Senthil
  Purushwalkam, Stefano Ermon, Caiming Xiong, Shafiq Joty, and Nikhil Naik.
\newblock {Diffusion Model Alignment Using Direct Preference Optimization}.
\newblock In \emph{IEEE/CVF Conference on Computer Vision and Pattern
  Recognition}, 2024.

\bibitem[Wang et~al.(2023)Wang, Lu, Wang, Bao, Li, Su, and
  Zhu]{wang2023prolificdreamer}
Zhengyi Wang, Cheng Lu, Yikai Wang, Fan Bao, Chongxuan Li, Hang Su, and Jun
  Zhu.
\newblock {ProlificDreamer: High-Fidelity and Diverse Text-to-3D Generation
  with Variational Score Distillation}.
\newblock In \emph{Advances in Neural Information Processing Systems}, 2023.

\bibitem[Wu et~al.(2026)Wu, Luo, Zheng, Zhang, and Lin]{wu2026diffinstruct}
Junyi Wu, Weijian Luo, Haoyang Zheng, Ruizhe Zhang, and Guang Lin.
\newblock {Diff-Instruct with Diffused Reward: Towards Principled One-Step
  Generator RL}.
\newblock In \emph{Advances in Neural Information Processing Systems}, 2026.

\bibitem[Wu et~al.(2023)Wu, Hao, Sun, Chen, Zhu, Zhao, and Li]{wu2023human}
Xiaoshi Wu, Yiming Hao, Keqiang Sun, Yixiong Chen, Feng Zhu, Rui Zhao, and
  Hongsheng Li.
\newblock {Human Preference Score v2: A Solid Benchmark for Evaluating Human
  Preferences of Text-to-Image Synthesis}.
\newblock \emph{arXiv preprint arXiv:2306.09341}, 2023.

\bibitem[Xu et~al.(2023)Xu, Liu, Wu, Tong, Li, Ding, Tang, and
  Dong]{xu2023imagereward}
Jiazheng Xu, Xiao Liu, Yuchen Wu, Yuxuan Tong, Qinkai Li, Ming Ding, Jie Tang,
  and Yuxiao Dong.
\newblock {ImageReward: Learning and Evaluating Human Preferences for
  Text-to-Image Generation}.
\newblock In \emph{Advances in Neural Information Processing Systems}, 2023.

\bibitem[Xue et~al.(2026)Xue, Ge, Zhang, Li, and Ma]{xue2026advantage}
Shuchen Xue, Chongjian Ge, Shilong Zhang, Yichen Li, and Zhi-Ming Ma.
\newblock {Advantage Weighted Matching: Aligning RL with Pretraining in
  Diffusion Models}.
\newblock In \emph{International Conference on Machine Learning}, 2026.

\bibitem[Xue et~al.(2025)Xue, Wu, Gao, Kong, Zhu, Chen, Liu, Liu, Guo, Huang,
  et~al.]{xue2025dancegrpo}
Zeyue Xue, Jie Wu, Yu Gao, Fangyuan Kong, Lingting Zhu, Mengzhao Chen, Zhiheng
  Liu, Wei Liu, Qiushan Guo, Weilin Huang, et~al.
\newblock {DanceGRPO: Unleashing GRPO on Visual Generation}.
\newblock \emph{arXiv preprint arXiv:2505.07818}, 2025.

\bibitem[Yin et~al.(2024{\natexlab{a}})Yin, Gharbi, Park, Zhang, Shechtman,
  Durand, and Freeman]{yin2024improved}
Tianwei Yin, Micha{\"e}l Gharbi, Taesung Park, Richard Zhang, Eli Shechtman,
  Fr{\'e}do Durand, and William~T. Freeman.
\newblock {Improved Distribution Matching Distillation for Fast Image
  Synthesis}.
\newblock In \emph{Advances in Neural Information Processing Systems},
  2024{\natexlab{a}}.

\bibitem[Yin et~al.(2024{\natexlab{b}})Yin, Gharbi, Zhang, Shechtman, Durand,
  Freeman, and Park]{yin2024onestep}
Tianwei Yin, Micha{\"e}l Gharbi, Richard Zhang, Eli Shechtman, Fr{\'e}do
  Durand, William~T. Freeman, and Taesung Park.
\newblock {One-Step Diffusion with Distribution Matching Distillation}.
\newblock In \emph{IEEE/CVF Conference on Computer Vision and Pattern
  Recognition}, 2024{\natexlab{b}}.

\bibitem[You et~al.(2025)You, Cai, Gu, Xue, and Dong]{you2025teaching}
Zhiyuan You, Xin Cai, Jinjin Gu, Tianfan Xue, and Chao Dong.
\newblock {Teaching Large Language Models to Regress Accurate Image Quality
  Scores Using Score Distribution}.
\newblock In \emph{IEEE/CVF Conference on Computer Vision and Pattern
  Recognition}, 2025.

\bibitem[Zheng et~al.(2026)Zheng, Chen, Ye, Wang, Zhang, Jiang, Su, Ermon, Zhu,
  and Liu]{zheng2026diffusionnft}
Kaiwen Zheng, Huayu Chen, Haotian Ye, Haoxiang Wang, Qinsheng Zhang, Kai Jiang,
  Hang Su, Stefano Ermon, Jun Zhu, and Ming-Yu Liu.
\newblock {DiffusionNFT: Online Diffusion Reinforcement with Forward Process}.
\newblock In \emph{International Conference on Learning Representations}, 2026.

\end{thebibliography}
}

\end{document}